\documentclass{article}
\usepackage{ijcai22}

\usepackage{times}
\usepackage{soul}
\usepackage{url}
\usepackage[hidelinks]{hyperref}
\usepackage[utf8]{inputenc}
\usepackage[small]{caption}
\usepackage{graphicx}
\usepackage{amsmath,amssymb}
\usepackage{amsthm}
\usepackage{booktabs}
\usepackage{algorithm}
\usepackage{algorithmic}
\usepackage{mathrsfs}

\title{Revealing the Excitation Causality between Climate and Political Violence \\via a Neural Forward-Intensity Poisson Process}

\author{
Schyler C. Sun$^1$
\and
Bailu Jin$^1$
\and
Zhuangkun Wei$^1$
\and
Weisi Guo$^{1,2}$\footnote{Corresponding Author.}
\affiliations
$^1$DARTeC, Cranfield University, UK\\
$^2$Alan Turing Institute, UK\\
\emails
\{Schyler.Sun,~Bailu.Jin,~Zhuangkun.Wei,~Weisi.Guo\}@cranfield.ac.uk,
wguo@turing.ac.uk
}

\begin{document}

\maketitle

\begin{abstract}
The causal mechanism between climate and political violence is fraught with complex mechanisms. Current quantitative causal models rely on one or more assumptions: (1) the climate drivers persistently generate conflict, (2) the causal mechanisms have a linear relationship with the conflict generation parameter, and/or (3) there is sufficient data to inform the prior distribution. Yet, we know conflict drivers often excite a social transformation process which leads to violence (e.g., drought forces agricultural producers to join urban militia), but further climate effects do not necessarily contribute to further violence. Therefore, not only is this bifurcation relationship highly non-linear, there is also often a lack of data to support prior assumptions for high resolution modeling.

Here, we aim to overcome the aforementioned causal modeling challenges by proposing a neural forward-intensity Poisson process (NFIPP) model. The NFIPP is designed to capture the potential non-linear causal mechanism in climate induced political violence, whilst being robust to sparse and timing-uncertain data. Our results span 20 recent years and reveal an excitation-based causal link between extreme climate events and political violence across diverse countries. Our climate-induced conflict model results are cross-validated against qualitative climate vulnerability indices. Furthermore, we label historical events that either improve or reduce our predictability gain, demonstrating the importance of domain expertise in informing interpretation.
\end{abstract}

\section{Introduction}
Armed violence is on the rise. Whilst the rate of major wars has decreased over the past few decades, the number of civil conflicts has doubled since the 1960s, and political violence (e.g., terrorism, sectarian violence) have become more frequent in the past ten years. Governments and the international community often have little warning of impending crises. Current practices in conflict risk interpretation rely on diplomatic reports and extrapolating from statistical data. Machine learning is poised to boost the power of these approaches \cite{Guo18Nature}.

Among the most worrying of the mooted impacts of climate change is an increase in conflict as people compete for diminishing resources such as arable land and water. Research over the past decade has established that climate variability may influence the risk of conflict \cite{Nature11,Nature19,review}. Despite broad expert agreement that there is a strong climate and conflict association \cite{ClimateChange}, quantitative modeling results are divergent or even contradictory \cite{Nature19} \cite{Peace14}. Modelling the impact of climate on conflict and predicting out-of-sample events remains challenging, especially when only a few variables are considered \cite{Ward13}. As such, there remain serious challenges in the assumptions laid out in current causal inference models. Based on the general idea of a causality test \cite{causality_test}, many papers and ourselves here propose to evaluate the causal link by testing whether additional extreme climate events would improve the performance of political violence prediction. 

\subsection{Background}
\label{transient}
Causal mechanisms that generate political violence often have an excitation mechanism \cite{Hawkes_Afghan}. The excitation effect that extreme climate events have on armed violence is often non-linear and bifurcates, e.g., extreme weather can cause agriculture community to shift to political violence and thereafter future climate changes are unlikely to have the same impact. Consequently, continuous functions of causal modeling that do not adjust for uncertain functional behaviour change is unlikely to be successful in complex societies \cite{IPCC12}. These continuous causal models include \cite{Causal18}: Granger causality and Predictability Improvement (PI), Conditional Mutual Information (CMI), and Convergent Cross Map (CCM) methods. 

For discrete excitation causal models, they are typically heavily parameterized especially on the time window, such that they are generally applicable only to a particular genre (e.g., IED attacks in a particular area \cite{Hawkes_NI}), and lack general understanding of the nature of climate vs. terrorism. Furthermore, terrorism data often suffer from sparsity, which is caused by a variety of factors ranging from under-reporting, lack of data in an emerging crisis area, to poor data quality. This can have an effect in non-linear likelihood estimators which is commonly used in current deep methods with linear loss function (e.g., LSTM). 

\subsubsection{Challenges of State-of-the-Art Models}
In literature, neural network modeling of conflict events attempt to account for the non-linear relationships in society, but this suffers from two challenges. For most countries, political violence data is sparse. As such, the prior distribution assumption of output (e.g., conflict events) becomes critical \cite{WDistance}. In neural networks, the output value is considered to be a random variable conforming to a Gaussian distribution with corresponding mean and variance related to the inputs \cite{DLGP}. However, the number of conflict events have been shown to follow a Poisson point process \cite{T_Poisson,point_process}. As such, the divergence between Gaussian and Poisson distribution can not be ignored when they have small mean values in sparse data. Hence, in modeling, we propose that any model should set \textit{Poisson processes as the prior} for the prediction model. Another challenge is predicting the precise time interval between attacks, where the state-of-the-art window size agnostic algorithm requires massive amounts of time series data \cite{window_size}, often unavailable to the conflict modeling genre. Therefore, in evaluation, we propose to set the \textit{posterior likelihood} of probabilistic prediction as the indicator of the model performance.

\subsection{Motivation for a Neural Forward-Intensity Model}
Our goal is to develop a general method that can self-configure to identify causal excitation signals and relate them to contextualised recent historical events. First of all, a specialized predictor is required for the excitation-based non-linear mechanism. Suppose a deep learning model has the inputs of the historical terrorism and climate data, and the target output of the attack forecasting. Then, there would exist two main challenges in this task, which raise additional requirements to both modeling and its evaluation. 

Poisson process has been recognized as a principled framework for modeling event data \cite{point_review}, where the intensity parameter is a function determined by relevant covariates. There are various derivatives of Poisson process, which focus on optimizing the linear form of intensity function with its covariates, e.g. Hawkes process \cite{Hawkes_NI,Hawkes_Afghan}, self-correcting process \cite{sc_Poisson}, reinforced Poisson processes \cite{reinforced_Poisson} and etc. In the most recent work \cite{aaai,nips}, the intensity function is modeled via a neural network \cite{IJCAI21}, which enables more non-linear covariates being captured.

\begin{figure}
    \center{\includegraphics[width=\linewidth]{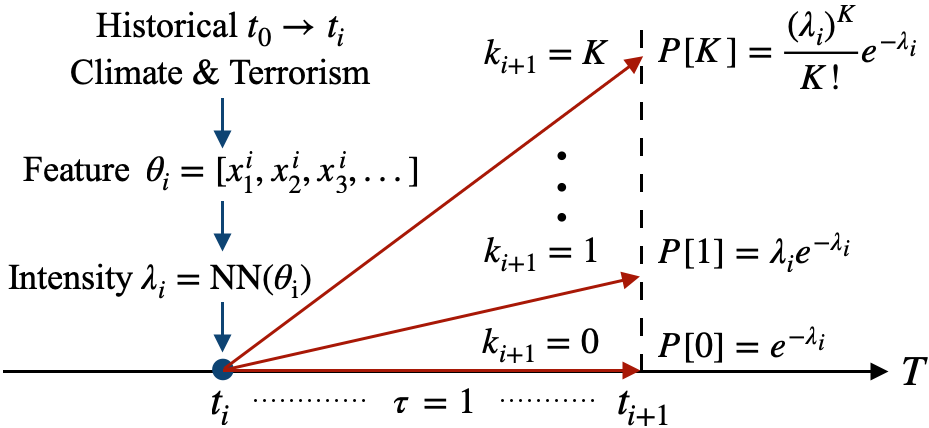}}
    \caption{The idea of neural forward-intensity Poisson forecasting at each time step is that:(1) Blue arrow: Intensity inference via a neural network (NN) using the extracted the features from historical covariates; (2) Red arrow: One step probabilistic forecasting based on the Poisson process with estimated intensity given.}
    \label{ForwardIntensity}
\end{figure}

As shown in Fig.\ref{ForwardIntensity}, in our proposed model, we adopt the NN-based intensity function to fit the non-linear mechanism between climate and conflict generation process. We make a one-step forward probabilistic forecasting via the Poisson process with the intensity value in each day. Our proposed neural forward-intensity Poisson process (NFIPP) model meets the requirements raised by the aforementioned challenges using the \textit{prior of Poisson} and the \textit{evaluation of likelihood}.

\subsection{Contribution and Novelty}
In this paper, we evaluate the excitation causality between climate and terrorism using the proposed neural forward-intensity Poisson process (NFIPP). Two major contributions of this paper are:

(i) In order to fit the non-linear mechanism in the sparse and timing-uncertain terrorist attack time series, we propose to embed a neural network based forward-intensity function in the Poisson process. This algorithm has been tested on a toy model and proven to have advantageous performance compared to conventional methods.

(ii) NFIPP is applied on the real terrorism and climate datasets for daily attack occurrence intensity inference. We reveal the excitation causal link from extreme climate events to terrorism via the posterior likelihood analysis. We find that the extreme climate events and other events (e.g., economy and politics) have the discrete excitation causal links to the terrorism rather than a continuous time causality, which require domain expertise in informing interpretation.

The remainder of this paper is organized as follows. In Section 2, we formulate the end-to-end learning framework for NFIPP model and verify its application scope with our assumption. In Section 3, we apply NFIPP to the real data and analyze the results with contextualised recent historical events. Section 4 concludes this paper and the ideas for future work.

\begin{figure*}
     \centering
     \includegraphics[width=1.0\linewidth]{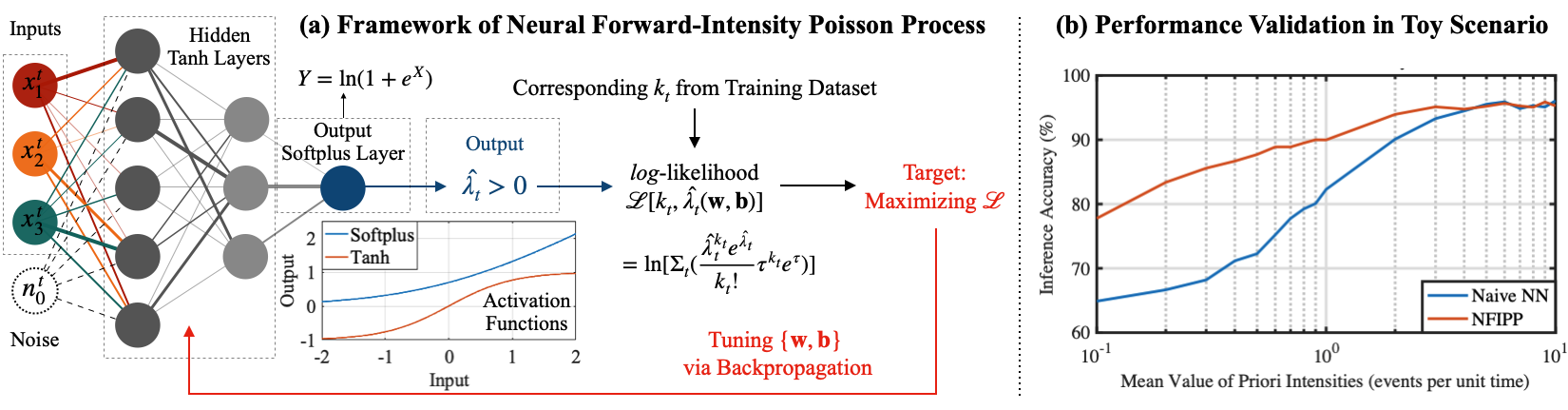}
     \caption{(a) The end-to-end learning framework of our proposed neural forward-intensity Poisson process (NFIPP) model. (b) The performance validation result of NFIPP in our supposed toy scenario, which shows its the advantage in low intensity range.}
     \label{Framework}
\end{figure*}

\section{Algorithm Formulation and Validation}
Our neural forward-intensity Poisson process (NFIPP) model is based on the assumption in function (\ref{Poisson}) that, the target time series $\mathbf{K} \in \mathbb{N}^{1\times T}$ (i.e., daily count $N$ of terrorist attacks) is the posterior observation $\{k_t\}^T_{t=1}$ from independent Poisson process with the corresponding forward-intensity $\lambda_t \in \mathbb{R}_{+}$ at each time slot $[t,t+\tau]$, while the intensity comes from a neural network (NN) which can be arbitrarily associated with the features $(x_1^t,x_2^t,x_3^t...)$ generated from former covariate series (e.g., terrorism and climate).
\begin{equation}
\label{Poisson}
\begin{split}
     P[N(t+\tau)&-N(t) = k]=\frac{(\lambda_t\tau)^k}{k!}e^{-\lambda_t\tau} \\
     \lambda_t &= \mathrm{NN}(x_1^t,x_2^t,x_3^t...)
\end{split}
\end{equation}
However, in reality the accurate intensity values are unknowable in most cases, hence we create a toy scenario to verify the performance and scope of our model before the implementation on the real-world data.

\subsection{End-to-End Learning Framework}
Fig.\ref{Framework}(a) demonstrates the end-to-end NFIPP learning framework. Taking the D-dimension feature vector at time $t$, $\mathbf{X_t} = [x_1^t;x_2^t;...;x_D^t] \in \mathbb{R}^{D \times 1}$, as the input, then NN is expected to has the output of the forward-intensity estimation $\hat{\lambda_t}$. Specifically, we introduce the “softplus” function $Y=\ln{(1+e^X)}$ in the last layer to ensure $\hat{\lambda_t}\ge 0$ meeting the positive constraint of Poisson process, while avoid vanishing gradient problem if "relu" function is used \cite{nips}.

With the outputs $\{\hat{\lambda_t}\}_{t=1}^T$ and the observation $\{k_t\}_{t=1}^T$ in the training data, the posterior log-likelihood $\Sigma\ln[P(k|\lambda)]$ can be derived as a function $\mathcal{L}(\mathbf{w},\mathbf{b})$ with the weight and bias parameters in the NN using function (\ref{Poisson}), which enable the training of $\{\mathbf{w},\mathbf{b}\}$ according to maximum log-likelihood via gradient back-propagation with learning rate $\eta$ as:
\begin{equation}
\left[
    \begin{matrix}
    \mathbf{w}_{i+1}\\
    \mathbf{b}_{i+1}\\
    \end{matrix} 
\right]
\leftarrow
\left[
    \begin{matrix}
    \mathbf{w}_{i}\\
    \mathbf{b}_{i}\\
    \end{matrix} 
\right]
+\eta
\left[
    \begin{matrix}
    \frac{\partial{\mathcal{L}(\mathbf{w_i},\mathbf{b_i})}}{\mathbf{w_i}}\\
    \frac{\partial{\mathcal{L}(\mathbf{w_i},\mathbf{b_i})}}{\mathbf{b_i}}\\
    \end{matrix} 
\right]
\end{equation}

\subsection{Model Scope Validation}
In order to verify the performance and scope of NFIPP model, we create a toy scenario with transparent intensity functions. In the toy scenario, suppose 
(i) $\mathbf{K} \in \mathbb{N}^{1\times T}$ are the posterior observations of corresponding Poisson process with the intensities $\mathbf{\Lambda} \in \mathbb{R}_{+}^{1\times T}$; 
(ii) Each element $\lambda_{t} \in \mathbf{\Lambda}$ is the function of potential features $\mathbf{x}^{t} \in [0,1]^D$, conditioned on $\lambda_{t} = \mathscr{F}(\mathbf{x}_{t}) \ge 0$. In the model validation, we set $T=1e3$, $D=4$, which are closed to the terrorism data training scale. For robustness test, we set $x_{4}^t$ to be an independent variable of the function $\mathscr{F}$, which act as a noise input. $\mathscr{F}$ randomly takes its form from a bunch of composite non-linear functions, including power, exponential, trigonometric and etc. (e.g. $\lambda = e^{x_1}\sin{(x_2)}+x_3^{\frac{1}{4}}$). In the test, we validate the average accuracy performance of our NFIPP over scenarios with mean of intensities varying from $10^{-1}$ to $10^{1}$, compared to conventional naive NN with the same neuron topology. With massive Monte-Carlo, the validation results are demonstrated in Fig.\ref{Framework}(b).

Both models are robust and without overfitting to the noise feature. Our NFIPP shows distinct advantage in prediction when the mean event occurrence rate is lower than 1 per unit time. This advantage comes from our improvement to the model priori of Poisson distribution. With the occurrence rate increasing to 4, Poisson distribution approximates to Gaussian distribution, where the advantage gradually fades away while the model performance keeps well. In reality for each country in last two decades, rarely do mean daily terrorist attack rate rises above 3 even in most dangerous countries (see Section 3). This means our NFIPP model would lead to significant improvement in the terrorism prediction.

\section{Implementation and Results}
As discussed in Section \ref{transient}, the excitation of climate in terrorism is non-linear and transient \cite{IPCC12}. In this section, we aim to reveal this kind of causal link a few representative countries with proposed NFIPP model, and analyze the potential underlying reasons backward from the results.

\subsection{Evaluation Metrics}
Based on the general idea of statistical causality test, we recognize a positive causality when additional historical climate data improve the performance of terrorism predictor comparing to historical terrorism only. In NFIPP, the prediction performance is evaluated by the posterior Poisson likelihood using the forward-intensity output given by predictor. Then the improvement can be calculated with the difference between the log-likelihood given by only historical terrorism data $\mathscr{L}_1$ and that the log-likelihood given by both historical terrorism and climate data $\mathscr{L}_2$. Since we only focus on the positive/negative effects and relative magnitude of improvement, for intuitive visualization, we define the likelihood gain with the difference normalized by a scalable $\tanh$ function:
\begin{equation}
    \mathrm{Likelihood \ Gain} = \tanh{[\alpha\times(\mathscr{L}_2-\mathscr{L}_1)]},
\end{equation} where $\alpha$ is an adaptive scaling parameter.

Besides, in order to give a comprehensive analysis, we introduce the prediction rate for reference. Since the model output is the forward-intensity $\lambda_t$ for time $[t,t+\tau]$ other than the an exact value, the count of terrorist attacks prediction within a period $[t_1,t_2]$ can only be calculated in a statistical way based on Poisson's property, i.e., $\hat{K} = \Sigma_{t_1}^{t_2} \lambda_t$. We define the prediction rate of the model over a specific time as:
\begin{equation}
    \mathrm{Prediction \ Rate} = \frac{\min(\hat{K},K)}{\max(\hat{K},K)},
\end{equation} which indicate the accuracy rate of matched prediction.

Furthermore, the overall impact of climate on the terrorism via the causal link is important for the cross-validation with results from political science \cite{social_science}. Here we use the ratio of additional correct predicted counts via climate data to the terrorism data as the index. For climate gain, we only consider the years that have positive likelihood gain, then the climate gain ratio is defined as:
\begin{equation}
    \mathrm{Climate\ Gain\ Ratio} = \frac{\max (0,\mathrm{TP \ Difference})}{\mathrm{Total\ Terrorism\ Counts}},
\end{equation} where the TP Difference is the true positive prediction difference between the counts from model with and without the climate data.

\subsection{Data Description}

\subsubsection{Data Sources}
The terrorism data come from Global Terrorism Database\footnote{https://www.start.umd.edu/gtd/} (GTD)  which is one of the most comprehensive global datasets on domestic and international terrorist attacks around the world. For extreme climate events, we consider two main aspects --- extreme rainfall and temperature events (e.g., floods and droughts). For flood and drought, the data come from EM-DAT \footnote{https://www.emdat.be}, which contains global essential core data on the occurrence of floods and droughts in countries. For extreme temperature, we consider the absolute anomaly value in the mean, lowest and highest temperature value, compared to standard baseline of local temperature, as the extreme temperature events. The corresponding data come from Berkeley Earth \footnote{http://berkeleyearth.org}, which supplies timely, impartial, and verified temperature data for countries. For cross validation, we use the \textit{country climate impact sensitivity index} introduced by ND-GAIN\footnote{https://gain.nd.edu/our-work/country-index/rankings/} \cite{social_science}. 

\subsubsection{Feature Pre-Processing}
In our implementation, each country are studied independently and the minimum time unit is one day. The target output for training is the count of terrorist attacks within country by day. The input feature attributes are formulated as follow: 
(i) Historical terrorism feature. We use the mean daily occurrence of attacks in past 15, 90 and 365 days to represent the trend of attack risks. 
(ii) Historical extreme climate event feature. We consider the climate influence in both short term (6 months) and long term (36 months), while 6 months is the average cycle period of crops, 36 months is the average period of El Niño and La Niña. For rainfall feature, we use the count of floods and droughts in past 6 and 36 months. For temperature feature, we take the sum of the absolute value of anomaly in past 6 and 36 months.

\subsubsection{Training and Inference}
In the experimental analysis, we perform a one-year step forward training and inference. In each step, six years' data are applied for training and thereafter the model gives the intensity prediction day by day in the next year. We mainly focus on the results from year 2000 to year 2019 whilst 6 years' more data from year 1994 are used for the initial training.

\subsection{Results and Discussion}
\begin{figure*}[htbp]
     \centering
     \includegraphics[width=0.99\linewidth]{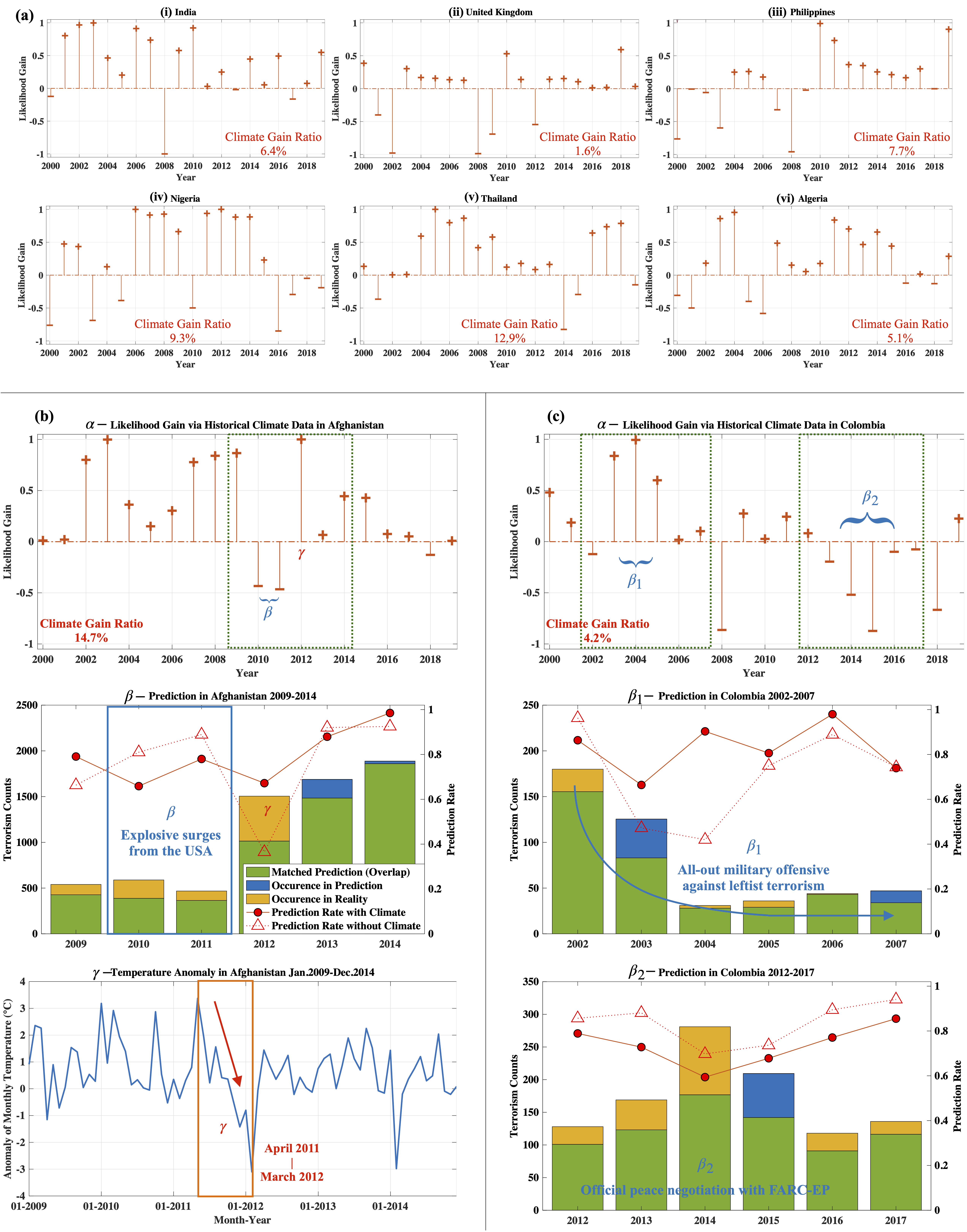}
     \caption{Experiment results. (a) General pattern analysis in representative countries; (b) Detailed analysis in Afghanistan; (c) Detailed analysis in Colombia.}
     \label{results}
\end{figure*}

We select a diverse range of 8 countries based on a combination of factors: total volume of political violence (1993-2019), geographic and ethnolinguistic diversity from each other, and genre of political violence. Fig.\ref{results} demonstrates the results for Afghanistan (AFG), Thailand (THA), Nigeria (NGA), India (IND), Philippines (PHL), Algeria (DZA), United Kingdom (GBR), Colombia (COL). The raw output from the NFIPP is daily based, but for simplicity of overall understanding, we take the period of one year in evaluation (e.g., likelihood gain).

\begin{figure}[b]
     \centering
     \includegraphics[width=1.0\linewidth]{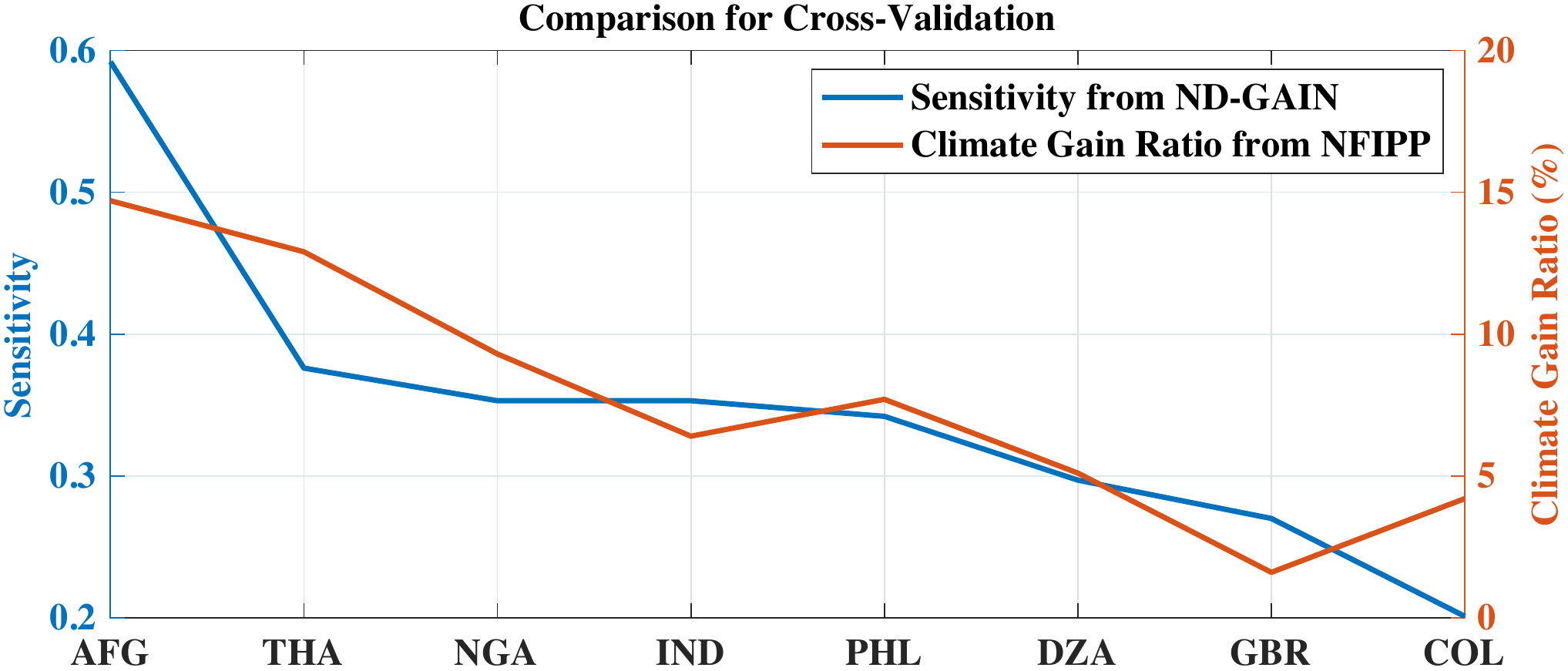}
     \caption{Cross-validation results across diverse countries and conflict situations.}
     \label{comparison}
\end{figure}

\subsubsection{Cross-Validation Against Climate Resilience Index}
The climate gain ratio indicate that to what extend can climate non-linearly excite the additional terrorism.  In order to cross-validate the results, we use the ND-GAIN country climate impact sensitivity index \cite{social_science}, which indicates the extent to a country depending upon a sector negatively affected by climate hazard and the proportion of the population particularly susceptible. This includes multiple sectors such as food, water, healthcare, ecosystem, human habitat, and infrastructure. The comparison is demonstrated in Fig.\ref{comparison}.The results from NFIPP and ND-GAIN show a strong similarity in trend. This allows us to verify that a country has a high causal link between extreme climate events (climate gain ratio) in our model also has a corresponding high sensitivity evaluated using more qualitative methods (e.g. ND-GAIN).

\subsubsection{General Analysis}
The positive likelihood gain represents the relative strength of the causal link from climate to terrorism, while the negative likelihood gain indicates that the attacks' mechanism is different from the climate induced pattern and have other causes. 

Firstly, we focus on the years with negative likelihood gain in Fig. \ref{results}a. From an economic and political point of view, most negative values can correspond to explanation in economic and political science. e.g., the 2008 financial crisis deeply affect the pattern in India, UK, and Philippines; the new allegiance of Boko Haram to the Islamic State of Iraq in Nigeria 2015 and the Royal Thai Army coup in Thailand 2014. These events acted as additional causal excitation to the terrorism mechanism and overwrote the causal link with climate. Secondly, if we trace backwards the climate data that have a spike positive likelihood gain via climate, the corresponding reasons can be found, e.g., Philippines reported over 60 floods and droughts during 2009-2011 which is much higher than usual years; Nigeria continuously suffered from anomaly high temperatures from 2002 to 2019.

\begin{table}[h]
    \centering
    \begin{tabular}{c|c|c}
    Category  &  Events Example     & Affected Countries\\
    \hline
    \hline
    Climate     &  Typhoon, Drought & ALL \\
    \hline
    Economy   & Financial Crisis  & IND, GBR, PHL, COL \\
    \hline
    Politics     & Coup, Surge & NGA, THA, AFG, COL \\
    \end{tabular}
    \caption{Example events that influence political violence.}
    \label{tab:events}
\end{table}

In Table \ref{tab:events}, we summarize the key recent historical events that we think have influenced the development of political violence in different countries. These conclusions are reinforced by various political and social science literature, but we acknowledge that further 
cross-domain expertise integration is the key to informing interpretation. In the next sub-section, we take two countries which have different geographic, ethnolinguistic, and political background, i.e. Afghanistan and Colombia, as the examples for a more detailed analysis.

\subsubsection{Detailed Analysis per Country}

(i) Afghanistan (Fig.\ref{results}b). Despite Afghanistan is involved in a protracted War on Terror (2001-2019), the causal link from climate to terrorism existed in most of these years and offer a significant 14.7\% gain ratio. Of interest is the likelihood gain fluctuation (Fig.\ref{results}b-$\alpha$) between 2009 and 2014. From Fig. \ref{results}b-$\beta$, we know that the total number of attacks in 2009-2011 are approximately the same, whilst the causal link break up in last two years. One possible reason was that the explosive surges from the USA (i.e. the troop level is doubled by the new president Barack Obama in those two years \cite{Surge1,Surge2}) just before the 2012 troop withdrawal, which resulted in the modification of terrorism pattern in 2010-2011. Besides, we notice that in 2012, the causal link reverse into a strong link and boost the prediction rate in that year. By looking back at the climate data, we found that, during the period of April 2011 to March 2021, the extreme temperature in Afghanistan experienced a cliff change from over-heating to under-cooling (Fig.(\ref{results})b-$\gamma$), while the events of drought and flood tripled compared to the past three years. By comprehensive analysis with Fig.(\ref{results})b, we believe that the there exist dominated excitation of climate in terrorism in Afghanistan.

(ii) Colombia (Fig. \ref{results}c). Climate events do not have great excitation on the terrorism in Colombia with the climate gain ratio of 4.2\%. However, potential causes can still be found if we take a closer look at the period of 2002-2007 ($\beta_1$) and 2012-2017 ($\beta_2$). Fig.(\ref{results})c-$\beta_{1}$ shows a sharp fall in the attack number from 2002, where the likelihood gain via climate become strong positive. One of the major reasons is that Álvaro Uribe was elected as the president of Colombia. He then led an all-out military offensive against leftist terrorism just following his 2002 election, which eliminated the political-based terrorist attacks. The remaining follow the unorganized attack pattern which cause by extreme climate events and drive a raise in the likelihood gain. Another notable period is 2012-2017 ($\beta_2$) where the likelihood gain gradually goes down and remain negative. Actually, from 2012 to 2016, the government of Colombia was on an official peace negotiation with the largest terrorism group FARC-RP. However, Fig.(\ref{results})c-$\beta_{2}$ shows an increase in attack number instead of decrease during the peace negotiation. By comprehensive analysis with Fig.(\ref{results})c, we believe that the politics and domestic terrorist groups are more dominated on the causal excitation terrorism in Colombia. 

The terrorism patterns are various from countries and time to time. In our analysis, the period of one year is taken for analysis. With further domain expertise, our results are day-by-day and adaptive for any time range analysis.

\section{Conclusion and Future Work}
The causal mechanism between climate and political violence is fraught with complex mechanisms. Current quantitative causal models largely rely on assumptions unsuitable for conflict mechanisms, namely: continuous and linear drivers, and sufficient data to model priors. Given that there is sufficient literature to show that drivers are often excitation based (e.g., a single event triggers a transformation but not thereafter), we overcome the aforementioned causal modeling challenges by proposing a neural forward-intensity Poisson process model. The model is designed to capture the potential non-linear excitation causal mechanisms in climate induced political violence, whilst being robust to sparse and timing uncertain data. Furthermore, the raw predictive output is a day-by-day Poisson process, improving temporal resolution over aggregated monthly/annual predictions. Our results span 20 recent years and reveal an excitation-based causal link between extreme climate events and political violence across diverse countries. Furthermore, we label historical events that either improve or reduce our predictability gain, demonstrating the importance of domain expertise in informing interpretation.

Our future work will try to take account of neighbourhood relations between countries to understand larger scale climate effects, political relations, and the diffusion of conflict. Here, we would expect a spatio-temporal point process \cite{Hawkes_Afghan} and a combined neural training of the diffusion intensity process. We hope the techniques developed help to forecast and limit future conflicts \cite{Guo18Nature}.

\bibliographystyle{named}
\bibliography{ijcai22}

\end{document}